# AI 3D CYBUG GAMING

Z. Ahmed

*Abstract*— **In this short paper I briefly discuss 3D war Game based on artificial intelligence concepts called AI WAR. Going in to the details, I present the importance of CAICL language and how this language is used in AI WAR. Moreover I also present a designed and implemented 3D War Cybug for AI WAR using CAICL and discus the implemented strategy to defeat its enemies during the game life.**

*Index Terms*— **AI, Game, Cybug**

## I. INTRODUCTION

3D War games have overcome the popularity of 2D games in international game markets. One of the most popular kinds of 3D games is WAR games, and in WAR games the most popular kind of game is artificially intelligent war games so called AI War Games. In most of the AI War games, the concept is to fight with enemies by using some weapons like guns, missiles, bombs, knife etc and complete the mission.

Where playing a 3D War game is very interesting there 3D War game development is a very complex task. It's always being a very hardworking task for the games designers and developers to produce a very good and strategic 3D game in best possible short time according to the need and demand of the market.

In this short research paper I am presenting the concept of 3D AI War game development in section II, then providing the information about CAICL in section III, then providing the information about Cybug behavior's implementation in section IV, then presenting own designed Ghazu in section V and in the end conclude discussion in section VI.

## II. 3D AI WAR GAME & ENGINE

AI War is a 3D game and an engine to design and develop war machines by using the concepts of artificial intelligence. The main concept of AI War is to design an artificial life based dynamic war scenario where every user's designed and developed war agent will be placed, which will then try to analyze, understand adopt provided circumstance, and using his provided weapons and designed expertise agent will fight with his enemy agents. During normal 3D War games a user is physically involved to play the game but in 3D AI War games a user is only there to design and develop his agent, but then the agent will use his own designed intelligence to play the game and the actual user will not be more than a viewer. In this case we can the user as a coach  who trains his player and during the match sits outside the ground  and watch his player playing with other players.

## III. CAICL

CAICL is the game development language used in AI War game programming [1].  In this AI War game every user's agent is called Cybug, CAICL language is used to develop Cybug consisting of logic commands variables and system variables. In AI War every Cybug can have many weapons like guns, missiles, grenades and energy discharges and shields, which can be used in war either for offense or defense purposes.

Moreover in AI War not only a Cybugs can fight with other one or more Cybugs but can also form a team and fight with other Cybugs to fight with team of other Cybugs, but the ultimate goal of each Cybug is to use his logical strategies and resources to win the war.

AI War engine provide a number of battlefield simulations to test the behavior of Cybug as individual as well as in collective Cybug war, moreover using AI War engine we can also customize a new war environment ourselves as well.

In AI War environment or battle field is called as Cybug Program. Its bit easy to design a simple Cybug but its bit very complex and difficult to design Cybug with learning behavior for example whenever a Cybug is loaded into a new Cybug has to fight against the other Cybug, moreover a map has provided to him but it's difficult to make Cybug aware of its complete environmental conditions.

But for this all Cybug and Cybug program development in AI War only the CAICL programming language is used.

## IV. CYBUG BEHAVIOR IMPLEMENTATION

In AI War there are five different types of behaviors which can be implemented in Cybug which are Sensing, Efficient coordination of movement and location, Dynamic allocation of resources, ability to model the opponents and Planning and problem solving [2].

### A. Sensing

Sensing is an intelligent behavior, can be implemented to track or detect enemy. By implementing sensing a Cybug can sense his enemy and obstacles by covering and good range of distance in four directions .i.e., *right, left, forward, backward*. To implement sensing behavior in Cybug there are different kinds of scans available in AI War engine like *Long Scan, GPS Scan, Scan forward, Scan right, Scan left* etc.

### B. Efficient coordination of movement and location

Efficient coordination of movement and location behavior is to make Cybug move in forward, backward, left and right in north, east, west, south directions.

Z. Ahmed is presently working as University Assistant with Mechanical Engineering Informatics and Virtual Product Development Division, Vienna University of Technology, Getreidemarkt 9/307 1060 Vienna Austria (phone: 004315880130726 - email: zeeshan.ahmed@tuwien.ac.at and zeeshan.ahmed@hotmail.de).

*C. Dynamic allocation of resources*

Dynamic allocation of resources is to make Cybug capable of dynamically allocating the different resources like fuel, weapons etc.

*D. Ability to model the opponent*

Ability to model the opponent is to make Cybug capable modeling the opponent or enemy.

*E. Planning and problem solving*

Planning and problem solving is to make Cybug fighting in battlefield with enemies and helping his friends (if he is fighting in team).

## V. GHAZU

Using the concepts of AI and provided options by AI War engine, we have designed and implemented a Cybug called Ghazu, capable of performing many sequences of actions e.g. scanning, utilizing and carrying resources and movements. Ghazu is capable of scanning enemies, flags, hurdles, damages and resources in the Cybug Program. Ghazu can also utilize and carry his resources .i.e., missiles, fuel, shields and flags. Moreover Ghazu is capable of moving right, left, forward and backward in Cybug Program. The strategy implemented in Ghazu is

1. Scan
   a. *if Found Enemy in range → attack*
   b. *if Found Flag in range → follow flag*
   c. *if Found Barrier in range → save*
   d. *if Found Mine in range → discharge energy*
   e. *if Found Fuel in range and needed → take fuel*
2. Damage
   a. *If Damaged Much & Enemy is in range → Suicide"*

The strategy implemented in Ghazu is very simple and divided in to two main steps .i.e., Scan and Damage.

In Scan five conditional instructions are implemented. At first if Ghazu gets any enemy in his range of action then fire and try to kill enemy, second if he finds flag in his range then go and try to get it before any other agent gets, third if he gets any barrier in his range then try to save his self from that, fourth if he find some mine in his way then try to discharge his energy to avoid mine attack and at last if he find fuel in his range and he is in need of fuel then go for it.

In Damage, if somehow Ghazu become injured badly and he is not in the condition to fight and he finds any enemy in his range to kill him, then he should blast his self and try to kill enemy by making a suicide attack.

## VI. CONCLUSION

In this short research paper we have briefly described AI War Game and engine, then going in to little bit more detail I have presented the concepts of CAICL AI War programming language and then presented a strategically designed and implemented Cybug.

## APPENDIX

GHAZU source code

```
name GHAZU
raise shield
Start:
long range scan
if scan found enemy then discharge energy
if scan found enemy then    lower shield
if scan found enemy then    launch missile
if scan found enemy then    raise shield
if scan found enemy then    turn right
if scan found enemy then    move forward
if scan found enemy then    turn left
goto Start

if scan found flag then move forward
if scan found flag then move forward
goto Start

if scan found flag then move forward
Bhagta:
turn right
move forward
if bump barrier then gosub Bhagta
goto start
Museebat:
generate random
if random is 1 then turn right
if random is 2 then turn left
if random is 3 then turn right
if random is 4 then turn left
move forward

if bump barrier then
goto Museebat
goto Start

if scan found mine then discharge energy
if fuel is < 99 goto then hide
if damage is >95 then goto Suiside

Suiside:
goto Start
lower shield
launch missile
self destruct
goto Start
```

ACKNOWLEDGEMENT

As the part of acknowledgement I would like to thanks Blekinge Institute of Technology, University of Blekinge to allow me to participate in their annual Cybug War Competition 2005.

[2]    Eager vs. Lazy Sensing, Reviewed 26 March 2009,
       <http://ai.eecs.umich.edu/cogarch2/prop/eager-lazy-sensing.html>